\pdfoutput=1

\documentclass[11pt]{article}

\usepackage{acl}

\usepackage{times}
\usepackage{latexsym}

\usepackage[T1]{fontenc}

\usepackage{microtype}

\usepackage{graphicx}

\usepackage{tabularx}
\usepackage{booktabs}
\usepackage{comment}
\usepackage{multirow}
\usepackage{url}
\usepackage{graphicx}
\usepackage{float}

\usepackage{makecell}

\usepackage{enumitem}

\title{Toward More Meaningful Resources for Lower-resourced Languages}

\author{
    Constantine Lignos \and Nolan Holley$^*$ \and Chester Palen-Michel$^*$ \and Jonne S{\"a}lev{\"a}$^*$ \\
    Michtom School of Computer Science\\
    Brandeis University\\
   \texttt{\{lignos,cpalenmichel,jonnesaleva\}@brandeis.edu}\\
   \texttt{nrh2@williams.edu}\\
}

\begin{document}
\maketitle
\def\thefootnote{*}\footnotetext{Denotes equal contribution. This paper was submitted to the ACL 2022 theme track ``Language Diversity: from Low-Resource to Endangered Languages'' and accepted to Findings of the ACL for ACL 2022.}\def\thefootnote{\arabic{footnote}}

\begin{abstract}
In this position paper, we describe our perspective on how meaningful resources for lower-resourced languages should be developed in connection with the speakers of those languages. We first examine two massively multilingual resources in detail. We explore the contents of the names stored in Wikidata for a few lower-resourced languages and find that many of them are not in fact in the languages they claim to be and require non-trivial effort to correct. We discuss quality issues present in WikiAnn and evaluate whether it is a useful supplement to hand annotated data. We then discuss the importance of creating annotation for lower-resourced languages in a thoughtful and ethical way that includes the languages' speakers as part of the development process. We conclude with recommended guidelines for resource development.
\end{abstract}

\section{Introduction}
Recent years have seen increased interest from the Association for Computational Linguistics (ACL) community in developing both models and datasets for what may be termed ``lower-resourced'' languages.
Advances in transfer learning and the increased availability of data and benchmarks in these languages have made it straightforward to create what appear to be high-performing models for these languages with little or no annotated data.

Yet despite the popularity and apparent effectiveness of these systems, the unique challenges and best practices for developing datasets and models for lower-resourced languages are rarely discussed alongside the system themselves.
In this position paper, through commentary on our experiences working with existing datasets and a discussion of current trends, we explore what resources can be most useful for the development of meaningful language technology for lower-resourced languages, advancing our perspective that \emph{open} data and models and a \emph{participatory} approach to research are the keys to progress.

Before we can elaborate on our perspective, we must address a terminological issue which creates a stumbling block when attempting to discuss resources and language technology for the languages of interest to us.
Throughout this paper, we will use the term \emph{lower-resourced language} to refer to languages that have received fewer resources---as measured in any numbers of dimensions such as models, datasets, papers, funding, etc.---than the most popularly-studied languages in the field of natural language processing.
We explicitly use the comparative \emph{lower} rather than \emph{low} to emphasize the continuum that exists across languages regarding the resources available for developing language technology.

We also acknowledge that whether a language is lower-resourced in a specific context may depend on what is available for the task at hand.
Due to singular efforts, a language that may have otherwise been underserved by the research community may have a rich set of resources for a single task like machine translation, but might not have annotation for other tasks.
For example, consider the Inuktitut language, which has a substantial amount of government-domain machine translation data \citep{joanis-etal-2020-nunavut} that enabled a shared task \citep{barrault-etal-2020-findings} creating many machine translation models, but has few labeled datasets for other tasks and is not included in any large multilingual language models we are aware of.

We take an open and intersectional perspective to what might be called a lower-resourced language, acknowledging that this designation is both imperfect and often the result of many contributing factors.
For example, many languages referred to in this way may be less-widely spoken, underserved by the research community and funding agencies, or used by marginalized or minoritized populations.
In summary, our use of the term \emph{lower-resourced} is intended to reflect a continuous, not categorical, status and one that is multi-dimensional, depending on task and context.

The goal of this paper is to argue for our perspective regarding what are the most effective ways to construct and use resources for building language technology for lower-resourced languages.
This perspective stems from our identity as researchers who are deeply committed to building impactful language technology, and we work with the speakers of lower-resourced languages to develop datasets and models that benefit their language communities.

This paper's contributions come at two levels.
At the more concrete level, we discuss particular issues related to using Wikidata and WikiAnn as sources of information about names, and highlight how automatic processes to take advantage of this information can go wrong when no human expertise is involved.
At a higher level, we discuss the problematic nature of developing language technology datasets and models with no or limited interaction with the population of speakers of the languages involved.

The structure of the paper is as follows.
We first begin a review of two popular resources used in NLP for lower-resourced languages and the impact that they have had on the field.
While we will discuss shortcomings of these resources, sometimes demonstrating them with experiments, the goal of this section is not to publish a critique of these resources, but rather to make other researchers aware of their shortcomings and limitations.
By acknowledging and understanding their limitations, we can better understand how to use them most appropriately and develop future resources that do not share the same limitations.

We then turn to the importance of annotation and dataset creation processes that meaningfully involve speakers of the languages under study.
We discuss open challenges for NLP for lower-resourced languages, and conclude with suggested guidelines for researchers performing research in this area.

\section{Wikidata: A source for name labels}

Wikidata\footnote{\url{https://www.wikidata.org}} is an open and collaboratively edited knowledge graph, hosted by the Wikimedia Foundation. 
The Wikidata graph consists of entity nodes connected by labeled edges that represent relations.
Each entity and relation is identified by a unique Wikidata identifier, e.g. Q4346375 (\textit{Association for Computational Linguistics}) and P361 (\texttt{part-of} relation).

While English labels  are typically used for the page titles on the Wikidata website, most entities have labels available in several languages, with the most well-edited entries having labels in hundreds of languages.
This makes Wikidata an appealing data source for constructing multilingual NLP resources related to entity names, as parallel names can in theory be trivially extracted from each entity.
For example, Wikidata could be used to harvest name lists for a named entity recognition (NER) system, or as a source of parallel names for translation or transliteration models.
In this section, we show Wikidata's promise for extracting names in lower-resourced languages as well as the data issues that arise in attempting to use it for this purpose.

\subsection{Name quality in lower-resourced languages}

Given the limited name-related annotation available for many lower-resourced languages, Wikidata is a promising source of information regarding entity names.
For example, previous work has used it as a source of data for multilingual name transliteration \citep{benites-etal-2020-translit,irvine2010transliterating}.
Specifically for lower-resourced languages, many approaches to NER and linking for the LORELEI program \citep{strassel-tracey-2016-lorelei} used Wikidata, Wikipedia, DBpedia, GeoNames, and similar resources to provide name lists relevant to the languages and regions for which systems were developed.

However, there are many caveats hidden in the data present in Wikidata and using the contents without scrutiny can be problematic. 
One such caveat is the mixing of languages and scripts occurring within the entity labels of a single language, especially lower-resourced ones.

Tigrinya, a Semitic language spoken in Africa by over 9 million speakers, is a particularly good example to explore, as it is written using the Ge'ez script and has only 539 entity labels in Wikidata. 
However, only 269 out of 539 labels are actually written in the Ge'ez script, with the rest being in Latin script.\footnote{We have confirmed with a native speaker of the language that this does not represent meaningful variation where some names may be borrowed in Latin script; they believed only Ge'ez script names should count as valid data.}
This problem is particularly pronounced among entities referring to persons, where only 36 entities are written in Ge'ez, and 245 in Latin script.
Out of all Tigrinya entity labels, nearly 50\% are identical to the English label.

While we have not done an exhaustive analysis, we believe that many other lower-resourced languages are affected by issues of this type.
Another example is Inuktitut, an Indigenous and historically minoritized language spoken in the Canadian Arctic by approximately 35,000--40,000 people, contains 25,222 entity labels in Wikidata, yet only 429 are written in the official Inuktitut syllabics.
Out of all Inuktitut labels, 97\% are identical to the corresponding English label, which suggests the entities are not only in the wrong script, but also the wrong language.

This vast number of Latin script labels relative to the ones written in Ge'ez script and Inuktitut syllabics suggests that labels are potentially being copied over from other languages, presumably as a result of bot activity. 
This script pollution may have adverse effects, particularly when the amount of data in the desired script is very small, not only on training models on the raw data but also on any heuristic filtering methods that try to, for example, filter out all entity labels not in the most common script for the language (which may end up being the incorrect script).

\subsection{Name copying}

Another minority language heavily impacted by likely copying is Asturian, a language spoken by 100,000--450,000 speakers in Spain.
Wikidata contains over 5 million entity labels for Asturian, with 97\% of them identical to the English label.
For comparison, 93.5\% out of 5.8 million Spanish entity labels are identical to English.

Overlap with English labels is not necessarily indicative of the labels being incorrect, as both Asturian and Spanish use the Latin alphabet and many named entities, particularly persons and organizations, can be written in the same way across languages.
However, the vast number of labels relative to the number of Asturian speakers (and proportionately, active Wikidata editors), and the extra-high level of English-matching suggests that labels are being copied from other languages.
This automated copying is widespread, so much so that Asturian ranks as the fourth largest language in Wikidata as measured by number of entity labels, following English, Dutch, and Spanish.

All in all, these examples show that extra care must be taken when harvesting multilingual data for lower-resourced languages in cases where it is possible that data may have been copied from a higher-resourced language.
This problem is most visible in cases where a language uses a non-Latin script, but is likely to exist for many other languages.
Failing to exercise caution may result in creating low-quality derived datasets that may do more harm than good.
For example, if a multilingual dataset contains a large number of incorrect copies of English in other languages, it may make tasks appear easier than they are because of trivial transfer from English.


\subsection{Summary}

In spite of the shortcomings in data quality, we still believe Wikidata may be a valuable resource for language technology development, provided that enough effort is invested in data cleaning and validation.
This process can take many forms depending on the application and could include, among other things, automated identification and filtering of languages and scripts, analysis of label copying from higher-resourced languages, and even analysis of who is making edits to Wikidata (for example, to identify automated edits).

Instead of shunning Wikidata, we encourage researchers to contribute to making it better for the global research community, and especially for lower-resourced languages for which Wikidata may be one of the only resources available.
We also encourage researchers to use Wikidata collaboratively with the speakers of lower-resourced languages who can provide guidance on the quality of resources derived from it and help with the process of removing incorrect information.

\section{WikiAnn}

We now turn our attention to a different Wikipedia-related resource, one derived from it.
WikiAnn \citep{pan-etal-2017-cross} is a dataset originally created for named entity tagging and linking of 282 languages present in Wikipedia. 
\citeauthor{pan-etal-2017-cross} generate ``silver-standard'' named entity annotations ``by transferring annotations from English to other languages through cross-lingual links and KB properties, refining annotations through self-training and topic selection, deriving language-specific morphology features from anchor links, and mining word translation pairs from cross-lingual links.''

\subsection{A ``silver'' standard for multilingual NER}

Since the WikiAnn dataset was created it has been used as a multilingual NER benchmark and it is included as part of the XTREME \citep{hu2020xtreme} massively multilingual multi-task benchmark.
It is not uncommon for WikiAnn to be mentioned as a multilingual NER benchmark with sometimes no mention of the fact that it is system output and not, in fact, annotation (despite the \emph{Ann} in the name).

We question the appropriateness of treating WikiAnn as a multilingual benchmark for NER. 
Even in a higher resourced language, the practice of evaluating a task on automatically derived data is sub-standard, hence the original authors referring to it as a ``silver standard.'' 
This kind of evaluation can only show how close one model comes to replicating the automated data collection process and does not reflect human performance.

Just examining the English data of WikiAnn reveals a number of entity names that would otherwise never be marked as names by a human. 
Strings of text such as \emph{Independently released}, \emph{If I were a boy}, and \emph{List of books written by teenagers} are annotated as organizations. 
\emph{I was glad}, \emph{the latter's studio}, and \emph{were promoted} are annotated as locations, and \emph{range has expanded}, \emph{a twelve-year-old passenger was found alive} and \emph{Artavasdes II, who served as} are tagged as person. 
While these examples are a small sample, we identified hundreds like these with either span issues, Wikipedia-specific entities like lists, incorrect entity types or entities that were simply not names. One might argue that this represents less than 1\% of the English WikiAnn data and is therefore noise.
However, human annotation does not typically have such mistakes and also has the benefit of being able to report inter annotator agreement so that researchers can better understand the difficulty of the task.

\subsection{Subsampled splits: Unnecessarily discarded data}

\begin{figure}[tb]
    \centering
    \includegraphics{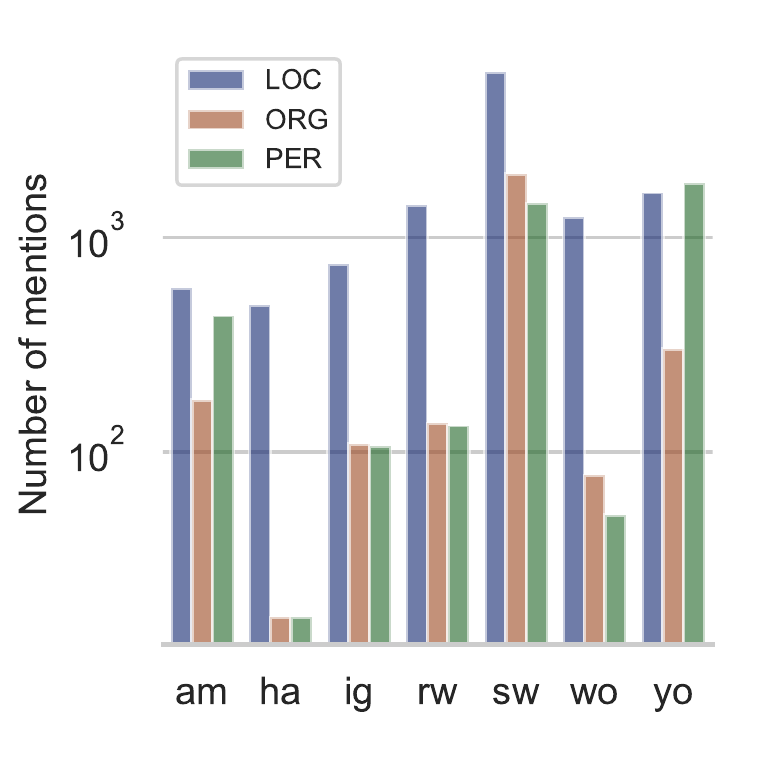}
    \caption{Counts of mentions in the full WikiAnn datasets in seven African languages that will be of interest in experiments in Section~\ref{section:masakhane-wikiann-expts}. A logarithmic scale is used so that all languages can be visualized.}
    \label{fig:wikiann-original}
\end{figure}

\begin{figure}[tb]
    \centering
    \includegraphics{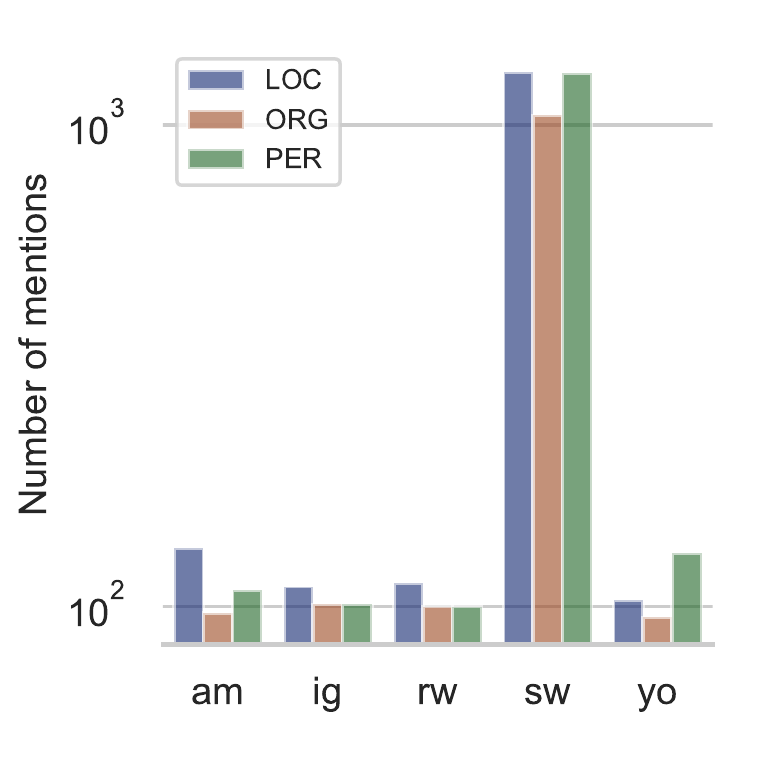}
    \caption{Counts of mentions in the stratified data selected by \citep{rahimi-etal-2019-massively}. A logarithmic scale is used so that all languages can be visualized.
    }
    \label{fig:wikiann-stratified}
\end{figure}

\begin{figure}[tb]
    \centering
    \includegraphics{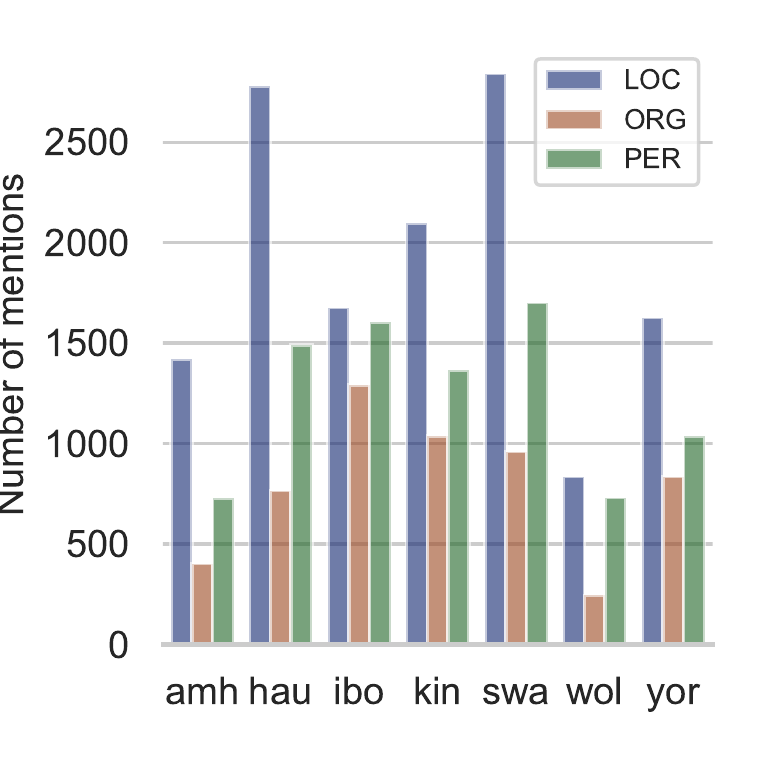}
    \caption{Counts of mentions in the MasakhaNER data for languages present in WikiAnn. The DATE tag was excluded for consistency with WikiAnn.}
    \label{fig:masakhane-entities}
\end{figure}

There can be some confusion as to which dataset is referred to by ``WikiAnn.'' 
The original WikiAnn paper \citep{pan-etal-2017-cross} contains data in 282 languages, but there is also another version derived from the original with only 176 languages \citep{rahimi-etal-2019-massively} with fixed data splits.
This derived version is the one that can be found in Hugging Face's Datasets.\footnote{\url{https://Hugging Face.co/datasets/wikiann}}
The original (unsplit) WikiAnn datasets can be found on Google Drive.\footnote{\url{https://drive.google.com/drive/folders/1bkK6ly_awxe9IgAKL16VVvCtjcYcDSw8}}

We describe the differences between the two versions of the dataset in detail because many may use the currently popular Hugging Face Datasets library \cite{lhoest-etal-2021-datasets} and never realize they have access to less data than is in the original.
Additionally, we highlight how the subsampled data no longer reflects a natural distribution of entity types and discards substantial amounts of data.

The original WikiAnn datasets are much larger than what was kept in the splits used by \citet{rahimi-etal-2019-massively}.
As can be seen in Figure~\ref{fig:wikiann-original}, in the original data the languages display an uneven distribution of mention types, with most languages having far more mentions of LOC than ORG or PER.\footnote{In all figures showing WikiAnn data, we use Wikimedia language codes rather than ISO 639-3 codes for consistency with the original data.}
This likely reflects a mix of the true distribution of named entities in the data and the fact that recall is typically highest for LOC entities.

\citet{rahimi-etal-2019-massively} used stratified sampling to select sentences for inclusion in their splits.
The process is as follows: first, the sentences in the dataset are categorized into three groups (LOC, ORG, PER) based on the entity type of the last mention in the sentence. 
The size of the smallest of these groups is defined to be the minimum count, and this number of sentences is taken from each group and added to a new list of sentences. 
This list is shuffled, and if it is large enough, will be used to make 10,000/10,000/10,000 splits. 
If not, the splits will be 1,000/1,000/1,000, and if there are not enough sentences for that, then the next step is 100/100/100. 
If there is not enough data for 100/100/100 splits, then the language will be skipped. 
However, this was not the process used to create splits for the 41 languages whose performance was examined in \cite{rahimi-etal-2019-massively}, though the authors provide information on those splits in the appendix.

As can be seen in Figure~\ref{fig:wikiann-stratified}, stratified sampling was largely successful in balancing the mention types across the splits.
However, a large amount of data is discarded by this process, and Hausa and Wolof are entirely thrown out because the entity type with the minimum count had too few mentions for this splitting method to be used.
Furthermore, the distribution of entity types is artificial, and does not match up with what might be the natural distribution of entity types. 
For comparison see the distribution of entity types from the MasakhaNER data \cite{TACL3129}, a human annotated NER dataset for 10 African languages, as shown in Figure~\ref{fig:masakhane-entities}.

\subsection{Is WikiAnn useful for languages with human annotation?}

\begin{table*}[tb]
\centering
\small
\begin{tabular}{l*{8}r}
\toprule
\thead[r]{Lang. \\ \\ \\ } & \thead[r]{Sentences \\ \\ \\ } & \thead[r]{Sentences ending \\ in a period \\ \\ } & \thead[r]{Sentences \\ consisting of a \\ single mention} & \thead[r]{Mentions \\ \\ \\ } & \thead[r]{Tokens \\ \\ \\ } & \thead[r]{Tokens inside \\ mentions (\%) \\ \\ } & \thead[r]{Average tokens \\ per mention \\ \\ } \\
\midrule
amh & 1,032   & 101  & 381 & 1,189 & 6,477 & 38.74 & 2.11 \\
hau & 489 & 176  & 223 & 517 & 3,650 & 16.58 & 1.17 \\ 
ibo & 937 & 284  & 568 & 968 & 6,387 & 27.59 & 1.82 \\
kin & 1,517 & 176  & 1,163 & 1,680 & 6,496 & 42.90 & 1.66 \\
swa & 7,589 & 2,353  & 3,113 & 9,315 & 43,085 & 57.36 & 2.65 \\
wol & 1,196 & 337  & 624 & 1,370 & 10,800 & 17.05 & 1.34 \\
yor & 3,438 & 396  & 2,285 & 3,716 & 18,319 & 62.44 & 3.08 \\
\bottomrule
\end{tabular}
\caption{Token, sentence, and mention statistics for all data in seven African languages contained in WikiAnn.}
\label{tab:wikiann-stats}
\end{table*}

For languages where there is no annotated NER data, WikiAnn is likely better than nothing. 
However, when annotated data is available for evaluation, can WikiAnn still be a useful resource? 
We conducted experiments by comparing models fine-tuned only on the MasakhaNER training data to those fine-tuned on the concatenation of WikiAnn with the MasakhaNER training data to see whether adding WikiAnn could improve performance by providing additional in-language training data.\footnote{The MasakhaNER dataset makes use of the DATE tag in addition to LOC, ORG, and PER which appear in both datasets. For our experiments all of the DATE mentions were removed and annotated as O.}
For comparison, we also experimented with fine-tuning using the concatenation of the training data across all languages in MasakhaNER. 

Qualitatively, the WikiAnn data differs greatly from any typical NER dataset annotated on news, such as MasakhaNER.
As can be seen in Table~\ref{tab:wikiann-stats}, the WikiAnn data is generally very dense in mentions, contains many ``sentences'' not ending in periods (which are likely not actually sentences at all), has a high number of ``sentences'' that consist of only a single mention.

For all experiments, training was done for 50 epochs. Training was done using the \texttt{train\_ner.py} script from the MasakhaNER GitHub.\footnote{\url{https://github.com/masakhane-io/masakhane-ner}}
Unless otherwise specified, each experiment was run with 10 different seeds.\footnote{The learning rate was 5$e$-5. The optimizer used was AdamW, with an epsilon value of 1$e$-8, and the scheduler was the script's default scheduler, called \texttt{get\_linear\_schedule\_with\_warmup}. The maximum sequence length was set to 164, and the training batch size was set to 32. Prediction was done with a maximum sequence length of 512, because smaller values led to a handful of tokens not receiving any predicted labels.}

For evaluation, the SeqScore\footnote{\url{https://github.com/bltlab/seqscore}} toolkit \citep{palen-michel-etal-2021-seqscore} was used, with the \texttt{conlleval} method of repairing invalid label sequences unless otherwise specified. 
Reported scores are the average of all training runs.\footnote{Throughout this paper, the Wilcoxon rank-sum test is used to evaluate statistical significance.}

\subsection{Fine-tuning with WikiAnn and MasakhaNER}

\begin{table}[tb]
\small
\centering
\begin{tabular}{l*{5}r}
\toprule
Lang. & MasakhaNER & +WikiAnn & $\Delta$ & $p$-value \\
\midrule
amh & \textbf{70.63} $\pm1.16$ & 69.02 $\pm1.72$ & 1.61 & \textbf{0.0191} \\
hau & \textbf{90.54} $\pm0.56$ & 90.03 $\pm0.58$ & 0.51 & 0.1041 \\
ibo & \textbf{86.37} $\pm0.83$ & 85.44 $\pm0.58$ & 0.93 & \textbf{0.0211} \\
kin & \textbf{73.89} $\pm2.12$ & 72.14 $\pm1.79$ & 1.75 & 0.0821 \\
swa & 87.90 $\pm0.64$ & \textbf{88.16} $\pm0.85$ & 0.26 & 0.4963 \\
wol & 68.12 $\pm1.38$ & \textbf{68.18} $\pm1.22$ & 0.06 & 0.8798 \\
yor & 78.23 $\pm0.99$ & \textbf{79.25} $\pm0.98$ & 1.02 & 0.0539 \\
\bottomrule
\end{tabular}
    \caption{Comparison of F1 scores between XLM-R fine-tuned using only MasakhaNER data and fine-tuned using MasakhaNER data and all available WikiAnn data in each language.}
    \label{tab:experiment2}
\end{table}

We experimented with fine-tuning XLM-R on the concatenation of the MasakhaNER training data with all available WikiAnn data in the corresponding language. 
The results are shown in Table~\ref{tab:experiment2}. 
On average, F1 decreased by 0.49 when adding WikiAnn to the training data.
Three languages (Swahili, Wolof, and Yoruba) show increases in performance, and while none are statistically significant, the improvement in Yoruba is marginally significant.

\subsection{Fine-tuning with all MasakhaNER languages}
\label{section:masakhane-wikiann-expts}

\begin{table}[tb]
\small
\centering
\begin{tabular}{l*{5}r}
\toprule
Lang. & Single lang. & All langs. & $\Delta$ & $p$-value \\
\midrule
amh & 71.19 $\pm1.20$ & \textbf{71.70} $\pm1.01$ & 0.51 & 0.3847 \\
hau & 89.78 $\pm0.41$ & \textbf{90.85} $\pm0.48$ & 1.07 & \textbf{0.0005} \\
ibo & 84.18 $\pm0.94$ & \textbf{85.72} $\pm0.60$ & 1.54 & \textbf{0.0024} \\
kin & 73.29 $\pm1.40$ & \textbf{74.67} $\pm0.79$ & 1.38 & \textbf{0.0283} \\
lug & 80.02 $\pm0.91$ & \textbf{80.88} $\pm0.73$ & 0.86 & \textbf{0.0413} \\
luo & 74.43 $\pm1.60$ & \textbf{77.19} $\pm1.17$ & 2.76 & \textbf{0.0015} \\
pcm & 87.89 $\pm0.72$ & \textbf{89.14} $\pm0.49$ & 1.25 & \textbf{0.0002} \\
swa & \textbf{87.43} $\pm0.56$ & 87.19 $\pm0.42$ & 0.24 & 0.1988 \\
wol & 64.74 $\pm1.82$ & \textbf{65.33} $\pm1.42$ & 0.59 & 0.4963 \\
yor & 77.63 $\pm1.17$ & \textbf{80.75} $\pm0.52$ & 3.12 & \textbf{0.0002} \\
\bottomrule
\end{tabular}
    \caption{Comparison of F1 scores between XLM-R fine-tuned on MasakhaNER data and fine-tuned on the concatenation of the MasakhaNER train splits for all languages.}
    \label{tab:masakhane-all-langs}
\end{table}

The inclusion of WikiAnn data in the training data offers mixed results at best.
Another option is to pool the training data across languages, creating a multilingual NER model trained on just the MasakhaNER data that is evaluated on each language's test set individually.\footnote{For this experiment, the DATE tag was left in the data, as consistency with WikiAnn data was not necessary.}
\citet{TACL3129} previously performed this experiment, but we replicate it here using our methodology, which includes statistical significance testing, a larger number of random seeds, and includes Amharic in this experiment even though it uses a different script than the other languages.

The results, shown in Table~\ref{tab:masakhane-all-langs}, are similar to those reported by \citep{TACL3129}, with all languages seeing improved performance except for Swahili.
Many of the improvements are statistically significant, showing that simply using more higher quality in-domain human-annotated data improves performance, while WikiAnn does not appear to help.

\subsection{Summary}

Many languages have no annotated data of the type provided by WikiAnn, and for those languages WikiAnn may prove useful as long as users are aware of its shortcomings.
Being an automatically created dataset, it contains noisy data, and it was constructed without input from speakers of almost all the languages contained in it.
Given that the core types (LOC, ORG, and PER) are intended to have the same definition across both WikiAnn and the MasakhaNER datasets, we predicted that augmenting MasakhaNER data with WikiAnn would improve performance.

But even when all available WikiAnn data is used, it does not improve the performance of models for the MasakhaNER data, and the simpler approach of simply pooling the MasakhaNER data across all languages produces better results.
This suggests that the noise level of the WikiAnn ``silver'' standard is very high, raising into doubt the validity of benchmarks which treat it as gold standard data.


\section{Human annotation: Still essential}

We have spent much of our paper describing many of the complexities of working with large-scale, Wiki-derived datasets, demonstrating that while they have some utility, they must be used with caution.
This caution stems from the fact that though the data contained in them originated from human contributions, in its final form in a resource, the data has been removed from its original quality checks.
For example, in Wikidata, names may be copied from one language to another \textit{en masse}, and in WikiAnn, the NER ``annotation'' is system output trained on relatively distant human supervision.

We have spent so much of our paper describing these shortcomings to demonstrate that there is no ``free lunch'' when it comes to avoiding human annotation or quality checking of datasets.
We believe that human annotation processes that are ultimately participatory---involving speakers of the languages as stakeholders and collaborators, not mere annotators for hire---like that of the MasakhaNER project which we have featured through the previous section and related projects \citep{nekoto-etal-2020-participatory,orife2020masakhane} are the most important direction for developing language technology for lower-resourced languages.

A discussion of efforts to annotate lower-resourced languages would not be complete without a discussion of the resources developed as part of the DARPA\footnote{Defense Advanced Research Projects Agency, a research agency that is part of the United States Department of Defense and funds a large proportion of US-based computer science research.} LORELEI (Low Resource Languages for Emergent Incidents) program.
The LORELEI program began in fall 2015, and a major thrust of the program was producing annotation for many lower-resourced languages \citep{strassel-tracey-2016-lorelei}.
The Linguistic Data Consortium (LDC) is in the process of releasing the 31 language packs developed as a part of the program, which have been available to the primarily U.S.-based groups funded by the program for years.
\citet{tracey-strassel-2020-basic} stated that they planned to release 1-2 packs per month in 2020.
However, as of November 2021, even though the research efforts of the program have largely concluded, only 7 language packs have been released to the general public: Akan, Amharic, Oromo, Somali, Tigrinya, Ukranian, and Vietnamese.
Each of these language packs is available for \$200 USD to non-members of the LDC.
While that is less expensive than typical LDC datasets, that cost can be prohibitively expensive for people who are speakers of the languages in the packs, who may live in countries with substantially lower wages and may not have the backing of a well-funded research lab. 

While it is unfortunate that so little data has been released to date and that the data is not freely available, the main contrast we would like to draw between LORELEI and efforts such as MasakhaNER is the involvement of the speakers of lower-resourced languages.
Speakers of the languages included in the LORELEI datasets did not have any significant involvement in the construction of the datasets beyond their role as annotators.
This is in no way unique to the LORELEI program; it is the status quo for annotation projects.

Returning to the issues we raise in our introduction, we want to highlight that a confluence of factors come together to make a language lower-resourced, among them often a marginalization and/or minoritization of its speakers.
We should consider whether it is ethical to have a paradigm in which the marginalized have no say in research that involves them, and in this case may not even be able to access the result of their work years after it is performed.
Developing resources for languages that have had fewer resources created for them to date poses a unique set of ethical challenges that differs from higher-resourced language work, and engaging language speaker in a participatory fashion can help mitigate the risk of harm.

\section{Challenges}
\label{section:challenges}

Before we conclude, we wish to highlight challenges that we believe should be addressed as a part of continuing to develop resources and models for lower-resourced languages.

\paragraph{Quality control for text resources.}

A popular way to gather multilingual data is through web crawling. 
These datasets include CCAligned, Multilingual C4 (mC4), OSCAR, ParaCrawl, WikiMatrix, and the aforementioned JW300.
However, as detailed in \citep{caswell2021quality}, there can be fairly serious quality issues when web-crawled data collection is not done carefully. 
Currently, when working with lower-resourced languages in large multilingual datasets, it is not a certainty (and sometimes, not even likely) that data is actually in the language that it claims to be.

\paragraph{Reducing reliance on religious text.}
Due to the large amount of translation of religious texts into lower-resourced languages by religious organizations in attempts to spread their message, religious materials are a common place to look for parallel or monolingual data.\footnote{From the perspective of decolonizing language technology \citep{bird-2020-decolonising}, these sources may be especially problematic as many of them were used as tools of colonization.}
While religious texts can be a convenient source of data due to their broad coverage of languages, it is important to be aware of potential biases, especially when the religion of the text is not the predominant religion of the speakers of the language---and thus may not match their norms---or when the target task could be affected by bias from the religious data. 

JW300 \citep{agic-vulic-2019-jw300} is a source of parallel data for over 300 languages with roughly 100,000 parallel sentences per language pair on average. 
The data was scraped from jw.org, which is the website of the Jehovah's Witnesses. 
Despite being sourced from a religious organization, it contains articles on a variety of topics translated into many languages.\footnote{At the time of writing, this resource is not available. The site distributing the dataset claimed that the dataset was freely available for non-commercial use, referred readers to jw.org's copyright at \url{https://www.jw.org/en/terms-of-use/}, and stated that ``for all practical purposes their custom terms of use are very closely aligned with the more well-known CC-BY-NC-SA license.''
However, recently the dataset has been taken down due to a copyright complaint until formal permission can be obtained. 
}
Inclusion of articles on a variety of topics does not fully prevent the potential for religious bias. 
As \citet{azunre2021contextual} demonstrate with a few masked sentence completion examples, a model trained on JW300 frequently produces completions with biblical names.
Although these types of completions are not grammatically incorrect, they are suggestive of a low level of generalization beyond religious data.


\section{Conclusion}

In closing, we want to refer to the state of affairs highlighted by \citet{bird-2020-decolonising}; some researchers are preoccupied with a data-centric view to the point of completely removing the need to involve speakers of the language in any part of the process.
Through our discussion of the shortcomings of Wikidata and WikiAnn for the specific purposes that we have evaluated them, we demonstrated the gaps that are created when the dataset creation process is divorced from the speakers of the language.
Our perspective on the process of dataset and model creation can be summarized through these guidelines we propose for future work on lower-resourced languages:
\begin{enumerate}[noitemsep]
    \item \label{guide:maximizecontact} Maximize interaction with and listening to the native speakers of languages included in resources you are developing.
    \item \label{guide:qa} When feasible, engage with speakers of included languages for quality control.
    \item \label{guide:lowquality} Consider the potential negative consequences of releasing datasets known to be of low-quality, as regardless of how you intend the resources to be used, they will likely be used for evaluation purposes.
    \item \label{guide:annotation} Prefer human annotation by speakers of the language to automatic processes, and release all human annotator decisions \citep{davani2021dealing, prabhakaran-etal-2021-releasing}.
\end{enumerate}

\bibliography{anthology,custom}
\bibliographystyle{acl_natbib}


\end{document}